\documentclass[11pt,a4paper]{article}
\usepackage[hyperref]{emnlp-ijcnlp-2019}
\usepackage{times}
\usepackage{soul}
\usepackage{latexsym}
\usepackage{amsmath}
\newcommand{\speedup}[0]{11x~}
\newcommand{\exact}[0]{10.73~} 
\newcommand{\ilptest}[0]{0.76~}
\usepackage{url}
\usepackage{graphicx}
\usepackage{booktabs}
\usepackage[ruled,vlined]{algorithm2e}
\usepackage{bm}
\usepackage{amsfonts}

\aclfinalcopy 

\newcommand{\hideforemnlp}[1]{}

\title{Query-focused Sentence Compression in Linear Time}

\author{Abram Handler {\normalfont and} Brendan O'Connor  \\
  College of Information and Computer Sciences \\
  University of Massachusetts Amherst \\ 
  {\tt \string{ahandler,brenocon\string}@cs.umass.edu} \\}

\date{}

\begin{document}
\maketitle
\begin{abstract}
Search applications often display shortened sentences which must contain certain query terms and must fit within the space constraints of a user interface. This work introduces a new transition-based sentence compression technique developed for such settings. Our query-focused method constructs length and lexically constrained compressions in linear time, by growing a subgraph in the dependency parse of a sentence. This theoretically efficient approach achieves an \speedup empirical speedup over baseline ILP methods, while better reconstructing gold constrained shortenings. Such speedups help query-focused applications, because users are measurably hindered by interface lags. Additionally, our technique does not require an ILP solver or a GPU.
\end{abstract}

\section{Introduction}\label{s:intro}

\begin{figure}[htb!]
\includegraphics[width=8.5cm]{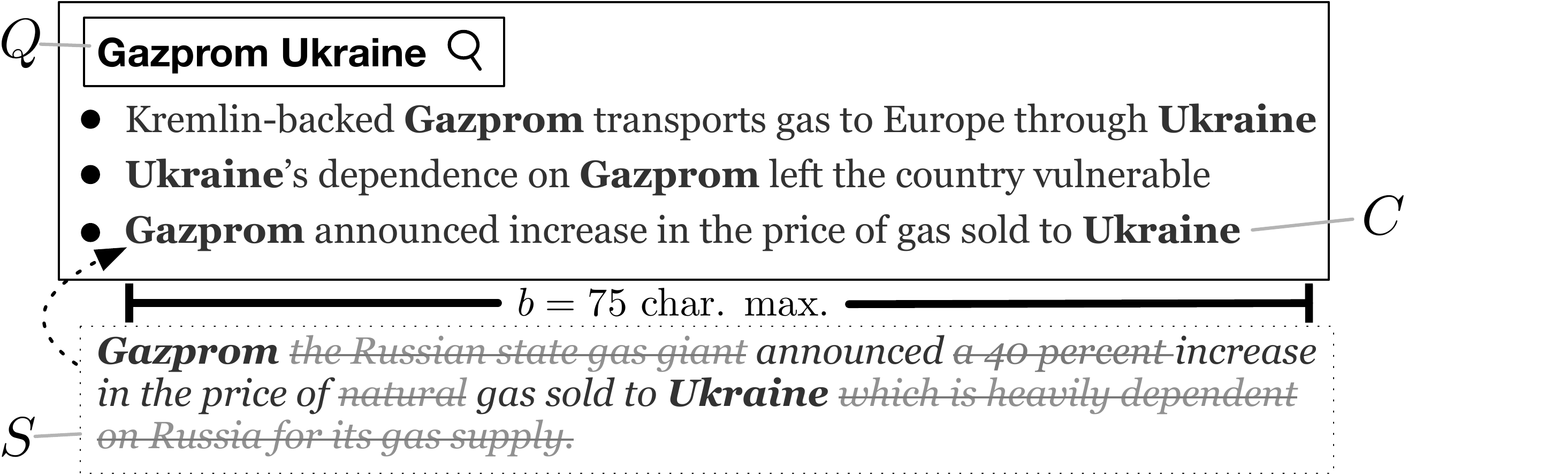}
\caption{A search user interface (boxed, top) returns a snippet consisting of three compressions which must contain a users' query $Q$ (bold) and must not exceed $b=$ 75 characters in length. The third compression $C$ was derived from source sentence $S$ (italics, bottom).}
\label{f:qf}
\end{figure}

Traditional study of extractive sentence compression seeks to create short, readable, single-sentence summaries which retain the most ``important'' information from source sentences. But search user interfaces often require compressions which must include a user's query terms and must not exceed some maximum length, permitted by screen space.  Figure \ref{f:qf} shows an example.

This study examines the English-language compression problem with such length and lexical requirements. In our constrained compression setting, a source sentence $S$ is shortened to a compression $C$ which (1) must include all tokens in a set of query terms $Q$ and (2) must be no longer than a maximum budgeted character length, $b \in \mathbb{Z}^{+}$. Formally, constrained compression maps $(S,Q,b) \rightarrow C$, such that $C$ respects $Q$ and $b$. We describe this task as query-focused compression because $Q$ places a hard requirement on words from $S$ which must be included in $C$.

Existing techniques are poorly suited to constrained compression. While methods based on integer linear programming (ILP) can trivially accommodate such length and lexical restrictions \cite{clarke2008global,filippova2013overcoming,Wang2017CanSH}, these approaches rely on slow third-party solvers to optimize an NP-hard integer linear programming objective, causing user wait time. An alternative LSTM tagging approach \cite{filippova2015sentence} does not allow practitioners to specify length or lexical constraints, and requires an expensive graphics processing unit (GPU) to achieve low runtime latency (access to GPUs is a barrier in fields like social science and journalism). These deficits prevent application of existing compression techniques in search user interfaces \cite{marchionini2006exploratory,hearst2009search}, where length, lexical and latency requirements are paramount. We thus present a new stateful method for query-focused compression.

Our approach is theoretically and empirically faster than ILP-based techniques, and more accurately reconstructs gold standard compressions.

\begin{table}[htb!]
\begin{tabular}{lcc}
Approach & Complexity & Constrained  \\ \hline
\textsc{ilp}       &   exponential    & yes     \\
LSTM tagger & linear              & no         \\   
\textbf{\textsc{vertex addition}} & \textbf{linear}     &      \textbf{yes}   
\end{tabular}
\caption{Our \textsc{vertex addition} technique (\S\ref{s:system}) constructs constrained compressions in linear time. Prior work (\S\ref{s:relatedwork}) has higher computational complexity (\textsc{ILP}) or does not respect hard constraints (LSTM tagger).} 
\label{t:algos}
\end{table}

\section{Related work}\label{s:relatedwork}

Extractive compression shortens a sentence by removing tokens, typically for summarization \cite{Knight2000StatisticsBasedS,clarke2008global,filippova2015sentence,Wang2017CanSH}.\footnote{Some methods compress via generation instead of deletion \cite{rush2015neural,mallinson18}. Our extractive method is intended for practical, interpretable and trustworthy search systems \cite{Chuang2012InterpretationAT}. Users might not trust abstractive summaries \cite{Zhang:2018:MSG:3290265.3274465}, particularly in cases with semantic error.} To our knowledge, this work is the first to consider extractive compression under hard length and lexical constraints.

We compare our \textsc{vertex addition} approach to ILP-based compression methods \cite{clarke2008global,filippova2013overcoming,Wang2017CanSH}, which shorten sentences using an integer linear programming objective. \textsc{ilp} methods can easily accommodate lexical and budget restrictions via additional optimization constraints, but require worst-case exponential computation.\footnote{ILPs are exponential in $|V|$ when selecting tokens \cite{clarke2008global} and exponential in $|E|$ when selecting edges \cite{filippova2015sentence}.} 

Finally, compression methods based on LSTM taggers \cite{filippova2015sentence} cannot currently enforce lexical or length requirements. Future work might address this limitation by applying and modifying constrained generation techniques \cite{D16-1140,N18-1119,D18-1443}.

\section{Compression via \textsc{vertex addition}}\label{s:system}

We present a new transition-based method for shortening sentences under lexical and length constraints, inspired by similar approaches in transition-based parsing \cite{nivre2003}. We describe our technique as \textsc{vertex addition} because it constructs a shortening by \textit{growing} a (possibly disconnected) subgraph in the dependency parse of a sentence, one vertex at a time. This approach can construct constrained compressions with a linear algorithm, leading to \speedup lower latency than ILP techniques (\S\ref{s:autoeval}). To our knowledge, our method is also the first to construct compressions by $\textit{adding}$ vertexes rather than \textit{pruning} subtrees in a parse \cite{Knight2000StatisticsBasedS,almeida2013fast,Filippova2015FastKS}. We assume a boolean relevance model: $S$ must contain $Q$. We leave more sophisticated relevance models for future work.

\subsection{Formal description}\label{s:formal}

\begin{figure}[h]
\includegraphics[width=8.2cm]{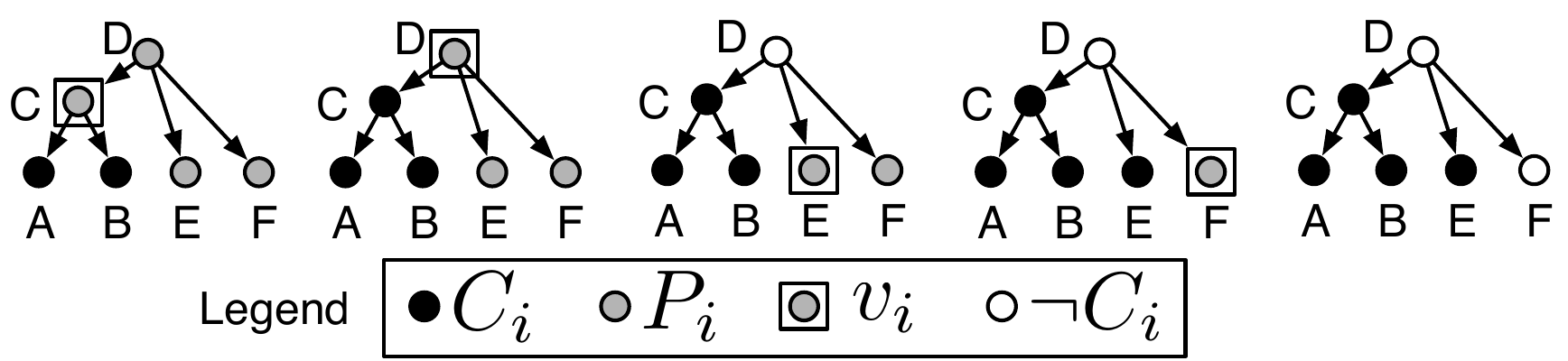}
\caption{A dependency parse of a sentence $S$, shown across five timesteps of \textsc{vertex addition} (from left to right). Each node in the parse is a vertex in $S$. Our stateful method produces the final compression $\{$A,C,B,E$\}$ (rightmost). At each timestep, each candidate $v_i$ is boxed; rejected candidates $\neg C_i$ are unshaded.}
\label{f:walkthru}
\end{figure}

\textsc{vertex addition} builds a compression by maintaining a state
$(C_i,P_i)$ where $C_i \subseteq S$ is a set of added candidates, $P_i  \subseteq S$ is a priority queue of vertexes, and $i$ indexes a timestep during compression. Figure \ref{f:walkthru} shows a step-by-step example. 

During initialization, we set $C_0 \gets Q$ and $P_0 \gets S \setminus Q$. Then, at each timestep, we pop some candidate $v_i =h(P_i)$ from the head of $P_i$ and evaluate $v_i$ for inclusion in $C_i$. (Neighbors of $C_i$ in $P_i$ get higher priority than non-neighbors; we break ties in left-to-right order, by sentence position). If we accept $v_i$, then $C_{i + 1} \gets C_i \cup v_i$; if not, $C_{i + 1} \gets C_i$. 
(Acceptance decisions are detailed in \S\ref{s:transition}.) 
We continue adding vertexes to $C$ until either $P_i$ is empty or $C_i$ is $b$ characters long.\footnote{We linearize $C$ by left-to-right vertex position in $S$, common for compression in English \cite{filippova2013overcoming}.} The appendix includes a formal algorithm. 

\textsc{vertex addition} is linear in the token length of $S$ because we pop and evaluate some vertex from $P_i$ at each timestep, after $P_0  \gets S \setminus Q$. Additionally, because (1) we never accept $v_i$ if the length of $C_i \cup v_i$ is more than $b$, and (2) we set $C_0 \gets Q$, our method respects $Q$ and $b$.

\section{Evaluation}\label{s:autoeval}

We observe the latency, readability and token-level F1 score of \textsc{vertex addition}, using a standard dataset \cite{filippova2013overcoming}.
We compare our method to an \textsc{ilp} baseline (\S\ref{s:relatedwork}) because ILP methods are the only known technique for constrained compression. All methods have similar compression ratios (shown in appendix), a well-known evaluation requirement \cite{napoles2011evaluating}. We evaluate the significance of differences between \textsc{vertex addition}$_{LR}$ and the \textsc{ilp} with bootstrap sampling \cite{D12-1091}. All differences are significant {\small $(p < .01)$}. 

\subsection{Constrained compression experiment}\label{s:constrained}

In order to evaluate different approaches to constrained compression, we require a dataset of sentences, constraints and known-good shortenings, which respect the constraints. This means we need tuples $(S, Q, b, C_g)$, where $C_g$ is a known-good compression of $S$ which respects $Q$ and $b$ (\S\ref{s:intro}).

To support large-scale automatic evaluation, we reinterpret a standard compression corpus \cite{filippova2013overcoming}
as a collection of input sentences and constrained compressions. The original dataset contains pairs of sentences $S$ and compressions $C_g$, generated using news headlines. For our experiment, we set $b$ equal to the character length of the gold compression $C_g$. We then sample a small number of nouns\footnote{1 to 3 nouns; cardinality chosen uniformly at random.} from $C_g$ to form a query set $Q$, approximating both the observed number of tokens and observed parts of speech in real-world search \cite{Jansen2000RealLR,Barr2008TheLS}. Sampled $Q$ include reasonable queries like ``police, Syracuse'', ``NHS'' and ``Hughes, manager, QPR''.

By sampling queries and defining budgets in this manner, we create {198,570} training tuples and {9949} test tuples, each of the form $(S,Q,b,C_g)$. \citet{filippova2013overcoming} define the train/test split. We re-tokenize, parse and tag with CoreNLP v3.8.0 \cite{corenlp}. We reserve 25,000 training tuples as a validation set. 

\subsection{Model: \textsc{ilp}}\label{s:ilp}

We compare our system to a baseline \textsc{ilp} method, presented in \citet{filippova2013overcoming}. This approach represents each edge in a syntax tree with a vector of real-valued features, then learns feature weights using a structured perceptron trained on a corpus of $(S,C_g)$ pairs.\footnote{Another ILP \cite{Wang2017CanSH} sets weights using a LSTM, achieving similar in-domain performance. This method requires a multi-stage computational process (i.e.\ run LSTM \textit{then} ILP) that is poorly-suited to query-focused settings, where low latency is crucial.} Learned weights are used to compute a global compression objective, subject to structural constraints which ensure $C$ is a valid tree. This baseline can easily perform constrained compression: at test time, we add optimization constraints specifying that $C$ must include $Q$, and not exceed length $b$.

To our knowledge, a public implementation of this method does not exist. We reimplement from scratch using \citet{gurobi}, achieving a test-time, token-level F1 score of \ilptest on the unconstrained compression task, lower than the result {\small (F1 = 84.3)} reported by the original authors. There are some important differences between our reimplementation and original approach (described in detail in the appendix). Since \textsc{vertex addition} requires $Q$ and $b$, we can only compare it to the ILP on the \textit{constrained} (rather than traditional, unconstrained) compression task.

\begin{table*}[htb!]
\centering
\begin{tabular}{lccc}
Approach & F1 & SLOR & $^{*}$Latency \\ \hline
\textsc{random} {\small (lower bound) }&{\small 0.653}&{\small 0.377}& {\small 0.5} \\
\textsc{ablated} {\small (edge only) }&{\small 0.827}&{\small 0.669}&{\small 3.7}\\
\textsc{vertex addition}$_{NN}$& {\small 0.873}& {\small 0.728}& {\small 2929.1} {\tiny  (CPU)} \\ \midrule 
\textsc{ilp}&{\small 0.852}& {\small 0.756}&{\small 44.0}\\
\textsc{vertex addition}$_{LR}$ & \small 0.881 & {\small 0.745}& {\small 4.1} \\
\end{tabular}
\caption{Test results for constrained compression. $^*$Latency is the geometric mean of observed runtimes (in milliseconds per sentence). \textsc{vertex addition}$_{LR}$ achieves the highest F1, and also runs \exact times faster than the \textsc{ilp}. Differences between all scores for \textsc{vertex addition}$_{LR}$ and \textsc{ilp} are significant {\tiny $(p < .01)$}.}
\label{t:results}
\end{table*}

\subsection{Models: \textsc{vertex addition}}\label{s:transition}

Vertex addition accepts or rejects some candidate vertex $v_i$ at each timestep $i$. 
We learn such decisions $y_i \in \{0,1\}$ using a corpus of tuples $(S,Q,b,C_g)$ (\S\ref{s:constrained}). Given such a tuple, we can always execute an oracle path shortening $S$ to $C_g$ by first initializing \textsc{vertex addition} and then, at each timestep: (1) choosing $v_i = h(P_i)$ and (2) adding $v_i$ to $C_i$ iff $v_i \in C_g$. We set $y_i=1$ if $v_i \in C_g$; we set $y_i=0$ if $v_i \notin C_g$. We then use decisions from oracle paths to train two models of inclusion decisions, ${p(y_i  = 1 | v_i, C_i, P_i, S)}$. At test time, we accept $v_i$ if $p(y_i > .5)$.

\textbf{Model One.} Our {\textsc{vertex addition}$_{NN}$ model broadly follows neural approaches to transition-based parsing (e.g.\ \citet{D14-1082}): we predict $y_i$ using a LSTM classifier with a standard max-pooling architecture \cite{D17-1070}, implemented with a common neural framework \cite{Gardner2017AllenNLP}. Our classifier maintains four vocabulary embeddings matrixes, corresponding to the four disjoint subsets $C_i \cup \neg C_i \cup P_i \cup \{v_i\} = V$. Each LSTM input vector $x_t$ comes from the appropriate embedding for $v_t \in V$, depending on the state of the compression system at timestep $i$. The appendix details network tuning and  optimization.

\textbf{Model Two.} Our \textsc{vertex addition}$_{LR}$ model uses binary logistic regression,\footnote{We implement with Python 3 using scikit-learn \cite{Pedregosa:2011:SML:1953048.2078195}. We tune the inverse regularization constant to $c=10$ via grid search over powers of ten, to optimize validation set F1.} with 3 classes of features.

\underline{Edge features} describe the properties of the edge $(u,v_i)$ between $v_i \in P_i$ and $u \in C_i$. We use the edge-based feature function from \citet{filippova2013overcoming}, described in detail in the appendix. This allows us to compare the performance of a vertex addition method based on local decisions with an ILP method that optimizes a global objective (\S \ref{s:ablated}), using the same feature set.

\underline{Stateful features} represent the relationship between $v_i$ and the compression $C_i$ at timestep $i$. Stateful features include information such as the position of $v_i$ in the sentence, relative to the right-most and left-most vertex in $C_i$, as well as history-based information such as the fraction of the character budget used so far. Such features allow the model to reason about which sort of $v_i$ should be added, given $Q$, $S$ and $C_i$.

\underline{Interaction features} are formed by crossing all stateful features with the type of the dependency edge governing $v_i$, as well as with indicators identifying if $u$ governs $v_i$, if $v_i$ governs $u$ or if there is no edge $(u,v_i)$ in the parse.

\subsection{Metrics: F1, Latency and SLOR}\label{s:f1}
We measure the token-level F1 score of each compression method against gold compressions in the test set. F1 is the standard automatic evaluation metric for extractive compression \cite{filippova2015sentence,Klerke2016ImprovingSC,Wang2017CanSH}. 

In addition to measuring F1, researchers often evaluate compression systems with human \textit{importance} and \textit{readability} judgements \cite{Knight2000StatisticsBasedS,filippova2015sentence}. In our setting $Q$ determines the ``important'' information from $S$, so importance evaluations are inappropriate. To check readability, we use the automated readability metric SLOR \cite{lau2015unsupervised}, which correlates with human judgements \cite{kannConl}. 

We evaluate theoretical gains from \textsc{vertex addition}  (Table \ref{t:algos}) by measuring empirical latency. For each compression method, we sample and compress $N=300,000$ sentences, and record the runtime (in milliseconds per sentence). We observe that runtimes are distributed log-normally (Figure \ref{t:times}), and we thus summarize each sample using the geometric mean. \textsc{ilp} and \textsc{vertex addition}$_{LR}$ share edge feature extraction code to support to fair comparison. We test \textsc{vertex addition}$_{NN}$ using a CPU: the method is too slow for use in search applications in areas without access to specialized hardware (Table \ref{t:results}). The appendix further details latency and SLOR experiments.

\subsection{Analysis:  \textsc{ablated} \& \textsc{random}}\label{s:ablated}
For comparison, we implement an \textsc{ablated} vertex addition method, which learns inclusion decisions using only edge features from \citet{filippova2013overcoming}. \textsc{ablated} has a lower F1 score than \textsc{ilp}, which uses the same edge-level information to optimize a global objective: adding stateful and interaction features (i.e.\ \textsc{vertex addition}$_{LR}$) improves F1 score. Nonetheless, strong performance from \textsc{ablated} hints that edge-level information alone (e.g.\ dependency type) can mostly guide acceptance decisions.

We also evaluate a \textsc{random} baseline, which accepts each $v_i$ randomly in proportion to $p(y_i = 1)$ across training data. \textsc{random} achieves reasonable F1 because (1) $C_0 = Q \in C_g$ and (2) F1 correlates with compression rate \cite{napoles2011evaluating}, and $b$ is set to the length of $C_g$.


\begin{figure*}[htb!]
\centering
\includegraphics[width=10cm]{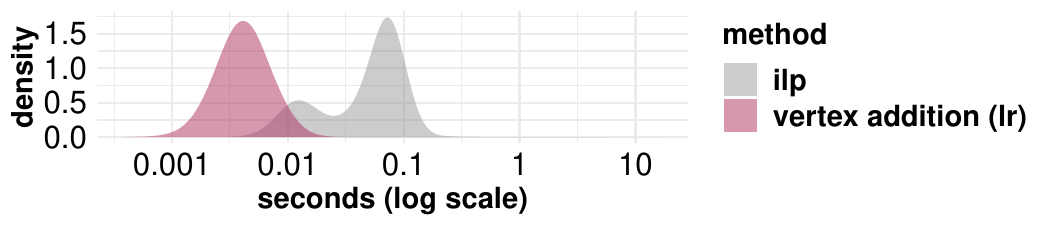}
\caption{Density plot of log transformed latencies for \textsc{Vertex Addition}$_{LR}$ (left) and \textsc{ilp} (right). Theoretical gains (Table \ref{t:algos}) create real speedups. The \textsc{ilp} shows greater runtime variance, possibly reflecting varying approaches from \citet{gurobi}.}
\label{t:times}
\end{figure*}

\section{Future work: practical compression}

This work presents a new method for fast query-focused sentence compression, motivated by the need for query-biased snippets in search engines \cite{tombros1998advantages,marchionini2006exploratory}. While our approach shows promise in simulated experiments, we expect that further work will be required before the method can be employed for practical, user-facing search.

To begin, both our technique and our evaluation ignore the conventions of search user interfaces, which typically display missing words using ellipses. This convention is important, because it allows snippet systems to transparently show users which words have been removed from a sentence. However, we observe that some well-formed compressions are difficult to read when displayed in this format. For instance the sentence ``Aristide quickly fled Haiti in September 1991'' can be shortened to the well-formed compression ``Aristide fled in 1991." But this compression does not read fluidly when using ellipses (``Aristide...fled...in...1991"). Human experiments aimed at enumerating the desirable and undesirable properties of compressions displayed in ellipse format (e.g.\ compressions should minimize number of ellipses?) could help guide user-focused snippet algorithms in future work. 

Our method also assumes access to a reliable, dependency parse, and ignores any latency penalties incurred from parsing. In practical settings, both assumptions are unreasonable. Like other NLP tools, dependency parsers often perform poorly on out-of-domain text \cite{bamman2017natural}, and users looking to quickly investigate a new corpus might not wish to wait for a parser. Faster approaches based on low-latency part-of-speech tagging, or more cautious approaches based on syntactic uncertainty \cite{keith-etal-2018-monte}, each offer exciting possibilities for additional research. 

Our approach also assumes that a user already knows a reasonable $b$ and reasonable $Q$ for a given sentence $S$.\footnote{Recall that we simulate $b$ and $Q$ based on the well-formed shortening $C_g$, see \S\ref{s:constrained}.} However, in some cases, there is no well-formed shortening of which respects the requirements. For instance, if $Q$=``Kennedy'' and $b$=15 there is no reasonable shortening for the toy sentence ``Kennedy kept running'', because the compressions ``Kennedy kept" and ``Kennedy running" are not well-formed. We look forward to investigating which $(Q,S,b)$ triples will never return well-formed compressions in later work. 

Finally, some shortened sentences will modify the meaning of a sentence, but we ignore this important complication in this initial study. In the future, we hope to apply ongoing research into textual entailment \cite{snli_bowman,Pavlick2016SoCalledNA,linzencompression} to develop semantically-informed approaches to the task.

\section{Conclusion}

We introduce a query-focused \textsc{vertex addition}$_{LR}$ method for search user interfaces, with much lower theoretical complexity (and empirical runtimes) than baseline techniques. In search applications, such gains are non-trivial: real users are measurably hindered by interface lags \cite{Nielsen,Liu2014TheEO}. We hope that our fast, query-focused method better enables snippet creation at the ``pace of human thought'' \cite{heerschei}.  

\section*{Acknowledgments}

Thanks to Javier Burroni and Nick Eubank for suggesting ways to optimize and measure performance of Python code. Thanks to Jeffrey Flanigan, Katie Keith and the UMass NLP reading group for feedback.
This work was partially supported by IIS-1814955.


\bibliography{abe}

\begin{thebibliography}{47}
\expandafter\ifx\csname natexlab\endcsname\relax\def\natexlab#1{#1}\fi

\bibitem[{Almeida and Martins(2013)}]{almeida2013fast}
Miguel Almeida and Andre Martins. 2013.
\newblock Fast and robust compressive summarization with dual decomposition and
  multi-task learning.
\newblock In \emph{ACL}.

\bibitem[{Bamman(2017)}]{bamman2017natural}
David Bamman. 2017.
\newblock Natural language processing for the long tail.
\newblock In \emph{Digital Humanities}.

\bibitem[{Barr et~al.(2008)Barr, Jones, and Regelson}]{Barr2008TheLS}
Cory Barr, Rosie Jones, and Moira Regelson. 2008.
\newblock The linguistic structure of english web-search queries.
\newblock In \emph{EMNLP}.

\bibitem[{Berg-Kirkpatrick et~al.(2012)Berg-Kirkpatrick, Burkett, and
  Klein}]{D12-1091}
Taylor Berg-Kirkpatrick, David Burkett, and Dan Klein. 2012.
\newblock An empirical investigation of statistical significance in {NLP}.
\newblock In \emph{EMNLP}.

\bibitem[{Bergstra and Bengio(2012)}]{Bergstra2012RandomSF}
James Bergstra and Yoshua Bengio. 2012.
\newblock Random search for hyper-parameter optimization.
\newblock \emph{Journal of Machine Learning Research}, 13:281--305.

\bibitem[{Bowman et~al.(2015)Bowman, Angeli, Potts, and Manning}]{snli_bowman}
Samuel~R. Bowman, Gabor Angeli, Christopher Potts, and Christopher~D. Manning.
  2015.
\newblock A large annotated corpus for learning natural language inference.
\newblock In \emph{EMNLP}.

\bibitem[{Briscoe et~al.(2006)Briscoe, Carroll, and
  Watson}]{briscoe-etal-2006-second}
Ted Briscoe, John Carroll, and Rebecca Watson. 2006.
\newblock The second release of the {RASP} system.
\newblock In \emph{Proceedings of the COLING/ACL 2006 Interactive Presentation
  Sessions}.

\bibitem[{Chen and Manning(2014)}]{D14-1082}
Danqi Chen and Christopher Manning. 2014.
\newblock A fast and accurate dependency parser using neural networks.
\newblock In \emph{EMNLP}.

\bibitem[{Chuang et~al.(2012)Chuang, Ramage, Manning, and
  Heer}]{Chuang2012InterpretationAT}
Jason Chuang, Daniel Ramage, Christopher~D. Manning, and Jeffrey Heer. 2012.
\newblock Interpretation and trust: {D}esigning model-driven visualizations for
  text analysis.
\newblock In \emph{CHI}.

\bibitem[{Clarke and Lapata(2008)}]{clarke2008global}
James Clarke and Mirella Lapata. 2008.
\newblock Global inference for sentence compression: An integer linear
  programming approach.
\newblock \emph{Journal of Artificial Intelligence Research}, 31:399--429.

\bibitem[{Conneau et~al.(2017)Conneau, Kiela, Schwenk, Barrault, and
  Bordes}]{D17-1070}
Alexis Conneau, Douwe Kiela, Holger Schwenk, Lo{\"i}c Barrault, and Antoine
  Bordes. 2017.
\newblock Supervised learning of universal sentence representations from
  natural language inference data.
\newblock In \emph{EMNLP}.

\bibitem[{Duchi et~al.(2011)Duchi, Hazan, and Singer}]{duchi2011adaptive}
John Duchi, Elad Hazan, and Yoram Singer. 2011.
\newblock Adaptive subgradient methods for online learning and stochastic
  optimization.
\newblock \emph{Journal of Machine Learning Research}, 12(Jul):2121--2159.

\bibitem[{Filippova and Alfonseca(2015)}]{Filippova2015FastKS}
Katja Filippova and Enrique Alfonseca. 2015.
\newblock Fast k-best sentence compression.
\newblock \emph{arXiv preprint arXiv:1510.08418}.

\bibitem[{Filippova et~al.(2015)Filippova, Alfonseca, Colmenares, Kaiser, and
  Vinyals}]{filippova2015sentence}
Katja Filippova, Enrique Alfonseca, Carlos~A Colmenares, Lukasz Kaiser, and
  Oriol Vinyals. 2015.
\newblock Sentence compression by deletion with \textsc{LSTM}s.
\newblock In \emph{EMNLP}.

\bibitem[{Filippova and Altun(2013)}]{filippova2013overcoming}
Katja Filippova and Yasemin Altun. 2013.
\newblock Overcoming the lack of parallel data in sentence compression.
\newblock In \emph{EMNLP}.
\newblock {h}ttps://github.com/google-research-datasets/sentence-compression.

\bibitem[{Filippova and Strube(2008)}]{filippova2008dependency}
Katja Filippova and Michael Strube. 2008.
\newblock Dependency tree based sentence compression.
\newblock In \emph{Proceedings of the Fifth International Natural Language
  Generation Conference}.

\bibitem[{Gardner et~al.(2017)Gardner, Grus, Neumann, Tafjord, Dasigi, Liu,
  Peters, Schmitz, and Zettlemoyer}]{Gardner2017AllenNLP}
Matt Gardner, Joel Grus, Mark Neumann, Oyvind Tafjord, Pradeep Dasigi,
  Nelson~F. Liu, Matthew Peters, Michael Schmitz, and Luke~S. Zettlemoyer.
  2017.
\newblock {AllenNLP}: A deep semantic natural language processing platform.

\bibitem[{Gehrmann et~al.(2018)Gehrmann, Deng, and Rush}]{D18-1443}
Sebastian Gehrmann, Yuntian Deng, and Alexander Rush. 2018.
\newblock Bottom-up abstractive summarization.
\newblock In \emph{EMNLP}.

\bibitem[{Gurobi~Optimization(2018)}]{gurobi}
LLC Gurobi~Optimization. 2018.
\newblock Gurobi optimizer reference manual (v8).

\bibitem[{Heafield(2011)}]{Heafield-kenlm}
Kenneth Heafield. 2011.
\newblock {KenLM:} faster and smaller language model queries.
\newblock In \emph{EMNLP: Sixth Workshop on Statistical Machine Translation}.

\bibitem[{Hearst(2009)}]{hearst2009search}
Marti Hearst. 2009.
\newblock \emph{Search user interfaces}.
\newblock Cambridge University Press, Cambridge New York.

\bibitem[{Heer and Shneiderman(2012)}]{heerschei}
Jeffrey Heer and Ben Shneiderman. 2012.
\newblock Interactive dynamics for visual analysis.
\newblock \emph{Queue}, 10(2).

\bibitem[{Jansen et~al.(2000)Jansen, Spink, and Saracevic}]{Jansen2000RealLR}
Bernard~J. Jansen, Amanda Spink, and Tefko Saracevic. 2000.
\newblock Real life, real users, and real needs: a study and analysis of user
  queries on the web.
\newblock \emph{Information Processing and Management}, 36:207--227.

\bibitem[{Kann et~al.(2018)Kann, Rothe, and Filippova}]{kannConl}
Katharina Kann, Sascha Rothe, and Katja Filippova. 2018.
\newblock {Sentence-Level Fluency Evaluation: References Help, But Can Be
  Spared!}
\newblock In \emph{{CoNLL 2018}}.

\bibitem[{Keith et~al.(2018)Keith, Blodgett, and
  O{'}Connor}]{keith-etal-2018-monte}
Katherine Keith, Su~Lin Blodgett, and Brendan O{'}Connor. 2018.
\newblock Monte {C}arlo syntax marginals for exploring and using dependency
  parses.
\newblock In \emph{NAACL}.

\bibitem[{Kikuchi et~al.(2016)Kikuchi, Neubig, Sasano, Takamura, and
  Okumura}]{D16-1140}
Yuta Kikuchi, Graham Neubig, Ryohei Sasano, Hiroya Takamura, and Manabu
  Okumura. 2016.
\newblock Controlling output length in neural encoder-decoders.
\newblock In \emph{EMNLP}.

\bibitem[{Klerke et~al.(2016)Klerke, Goldberg, and
  S{\o}gaard}]{Klerke2016ImprovingSC}
Sigrid Klerke, Yoav Goldberg, and Anders S{\o}gaard. 2016.
\newblock Improving sentence compression by learning to predict gaze.
\newblock In \emph{NAACL}.

\bibitem[{Knight and Marcu(2000)}]{Knight2000StatisticsBasedS}
Kevin Knight and Daniel Marcu. 2000.
\newblock Statistics-based summarization - step one: Sentence compression.
\newblock In \emph{AAAI}.

\bibitem[{Lau et~al.(2015)Lau, Clark, and Lappin}]{lau2015unsupervised}
Jey~Han Lau, Alexander Clark, and Shalom Lappin. 2015.
\newblock Unsupervised prediction of acceptability judgements.
\newblock In \emph{ACL}.

\bibitem[{Liu and Heer(2014)}]{Liu2014TheEO}
Zhicheng Liu and Jeffrey Heer. 2014.
\newblock The effects of interactive latency on exploratory visual analysis.
\newblock \emph{IEEE Transactions on Visualization and Computer Graphics},
  20:2122--2131.

\bibitem[{Mallinson et~al.(2018)Mallinson, Sennrich, and Lapata}]{mallinson18}
Jonathan Mallinson, Rico Sennrich, and Mirella Lapata. 2018.
\newblock {Sentence Compression for Arbitrary Languages via Multilingual
  Pivoting}.
\newblock In \emph{{EMNLP}}.

\bibitem[{Manning et~al.(2014)Manning, Surdeanu, Bauer, Finkel, Bethard, and
  McClosky}]{corenlp}
Christopher~D. Manning, Mihai Surdeanu, John Bauer, Jenny Finkel, Steven~J.
  Bethard, and David McClosky. 2014.
\newblock The {Stanford} {CoreNLP} natural language processing toolkit.
\newblock In \emph{ACL System Demonstrations}.

\bibitem[{Marchionini(2006)}]{marchionini2006exploratory}
Gary Marchionini. 2006.
\newblock Exploratory search: From finding to understanding.
\newblock \emph{Commun. ACM}, 49(4).

\bibitem[{de~Marneffe et~al.(2006)de~Marneffe, MacCartney, and
  Manning}]{Marneffe2006GeneratingTD}
Marie-Catherine de~Marneffe, Bill MacCartney, and Christopher~D. Manning. 2006.
\newblock Generating typed dependency parses from phrase structure parses.
\newblock In \emph{LREC}.

\bibitem[{McCoy and Linzen(2018)}]{linzencompression}
Thomas~R. McCoy and Tal Linzen. 2018.
\newblock Non-entailed subsequences as a challenge for natural language
  inference.
\newblock In \emph{Proceedings of the Society for Computation in Linguistics}.

\bibitem[{Napoles et~al.(2011)Napoles, Van~Durme, and
  Callison-Burch}]{napoles2011evaluating}
Courtney Napoles, Benjamin Van~Durme, and Chris Callison-Burch. 2011.
\newblock Evaluating sentence compression: Pitfalls and suggested remedies.
\newblock In \emph{Proceedings of the Workshop on Monolingual Text-To-Text
  Generation}.

\bibitem[{Nielsen(1993)}]{Nielsen}
Jakob Nielsen. 1993.
\newblock \emph{Usability Engineering}.
\newblock Morgan Kaufmann Publishers Inc., San Francisco, CA, USA.

\bibitem[{Nivre(2003)}]{nivre2003}
Joakim Nivre. 2003.
\newblock An efficient algorithm for projective dependency parsing.
\newblock In \emph{International Conference on Parsing Technologies}.

\bibitem[{Nivre et~al.(2016)Nivre, de~Marneffe, Ginter, Goldberg, Hajic,
  Manning, McDonald, Petrov, Pyysalo, Silveira, Tsarfaty, and
  Zeman}]{Nivre2016UniversalDV}
Joakim Nivre, Marie-Catherine de~Marneffe, Filip Ginter, Yoav Goldberg, Jan
  Hajic, Christopher~D. Manning, Ryan~T. McDonald, Slav Petrov, Sampo Pyysalo,
  Natalia Silveira, Reut Tsarfaty, and Daniel Zeman. 2016.
\newblock Universal {D}ependencies v1: A multilingual treebank collection.
\newblock In \emph{LREC}.

\bibitem[{Pavlick and Callison-Burch(2016)}]{Pavlick2016SoCalledNA}
Ellie Pavlick and Chris Callison-Burch. 2016.
\newblock So-called non-subsective adjectives.
\newblock In \emph{*SEM}.

\bibitem[{Pedregosa et~al.(2011)Pedregosa, Varoquaux, Gramfort, Michel,
  Thirion, Grisel, Blondel, Prettenhofer, Weiss, Dubourg, Vanderplas, Passos,
  Cournapeau, Brucher, Perrot, and
  Duchesnay}]{Pedregosa:2011:SML:1953048.2078195}
Fabian Pedregosa, Ga\"{e}l Varoquaux, Alexandre Gramfort, Vincent Michel,
  Bertrand Thirion, Olivier Grisel, Mathieu Blondel, Peter Prettenhofer, Ron
  Weiss, Vincent Dubourg, Jake Vanderplas, Alexandre Passos, David Cournapeau,
  Matthieu Brucher, Matthieu Perrot, and \'{E}douard Duchesnay. 2011.
\newblock Scikit-learn: Machine learning in python.
\newblock \emph{Journal of Machine Learning Research}, 12:2825--2830.

\bibitem[{Post and Vilar(2018)}]{N18-1119}
Matt Post and David Vilar. 2018.
\newblock Fast lexically constrained decoding with dynamic beam allocation for
  neural machine translation.
\newblock In \emph{NAACL}.

\bibitem[{Rush et~al.(2015)Rush, Chopra, and Weston}]{rush2015neural}
Alexander~M. Rush, Sumit Chopra, and Jason Weston. 2015.
\newblock A neural attention model for abstractive sentence summarization.
\newblock In \emph{EMNLP}.

\bibitem[{Schuster and Manning(2016)}]{Schuster2016EnhancedEU}
Sebastian Schuster and Christopher~D. Manning. 2016.
\newblock Enhanced english universal dependencies: An improved representation
  for natural language understanding tasks.
\newblock In \emph{LREC}.

\bibitem[{Tombros and Sanderson(1998)}]{tombros1998advantages}
Anastasios Tombros and Mark Sanderson. 1998.
\newblock Advantages of query biased summaries in information retrieval.
\newblock In \emph{SIGIR}.

\bibitem[{Wang et~al.(2017)Wang, Jiang, Chieu, Ong, Song, and
  Liao}]{Wang2017CanSH}
Liangguo Wang, Jing Jiang, Hai~Leong Chieu, Chen~Hui Ong, Dandan Song, and
  Lejian Liao. 2017.
\newblock Can syntax help? {I}mproving an {LSTM}-based sentence compression
  model for new domains.
\newblock In \emph{ACL}.

\bibitem[{Zhang and Cranshaw(2018)}]{Zhang:2018:MSG:3290265.3274465}
Amy~X. Zhang and Justin Cranshaw. 2018.
\newblock Making sense of group chat through collaborative tagging and
  summarization.
\newblock \emph{Proceedings of the ACM on Human-Computer Interaction}, 2(CSCW).

\end{thebibliography}
\bibliographystyle{acl_natbib}

\appendix
\appendix

\section{Appendix}

\subsection{Algorithm}
We formally present the \textsc{vertex addition} compression algorithm, using notation defined in \S{\ref{s:formal}}. $\ell$ linearizes a vertex set, based on left-to-right position in $S$. $|P|$ indicates the number of tokens in the priority queue.

\begin{algorithm}[]
\SetKwInOut{Input}{input}
\SetAlgoLined
\Input{$s=(V,E)$, $Q \subseteq V$, $b \in \mathbb{R^+}$}
 $ C \gets Q;  P \gets V \setminus Q$; \\
 \While{ $\ell(C) < b $ and $ |P| > 0 $   }{
  $v \gets \text{pop}(P)$; \\
  \If{$p(y=1) > .5$ and $\ell(C \cup \{v\}) \leq b$}{$C \gets C \cup \{v\}$}
 }
\KwRet{$\ell(C)$}
 \caption{\textsc{vertex addition}}
\end{algorithm}\label{a:algo}

\subsection{Neural network tuning and optimization}

We learn network parameters for \textsc{vertex addition}$_{NN}$ by minimizing cross-entropy loss against oracle decisions $y_i$. We optimize with \textsc{AdaGrad} \cite{duchi2011adaptive}. We learn input embeddings after initializing randomly. The hyperparameters of our network and training procedure are: the learning rate, the dimensionality of input embeddings, the weight decay parameter, the batch size, and the hidden state size of the LSTM. We tune via random search \cite{Bergstra2012RandomSF}, selecting parameters which achieve highest accuracy in predicting oracle decisions for the validation set. We train for 15 epochs, and we use parameters from the best-performing epoch (by validation accuracy) at test time.

\begin{table}[htb!]
\centering
\begin{tabular}{ll}
Learning rate & 0.025 \\ 
Embedding dim. &  315 \\
Weight decay   & 1.88 $ \times 10^{-9}$ \\
Hidden dim. & 158  \\
Batch size & 135 \\
\end{tabular}
\caption{Hyperparameters for \textsc{vertex addition}$_{NN}$}\label{t:params}
\end{table}

\subsection{Reimplementation of \citet{filippova2013overcoming}}

In this work, we reimplement the method of \citet{filippova2013overcoming}, who in turn implement a method partially described in \citet{filippova2008dependency}.  There are inevitable discrepancies between our implementation and the methods described in these two prior papers.

\begin{enumerate}
\item{Where the original authors train on only 100,000 sentences, we learn weights with the full training set to compare fairly with \textsc{vertex addition} (each model trains on the full training set).}
\item{We use \citet{gurobi} (v8) to solve the liner program. \citet{filippova2008dependency} report using LPsolve.\footnote{\url{http://
sourceforge.net/projects/lpsolve}}}
\item{We implement with the common Universal Dependencies (UD, v1) framework \cite{Nivre2016UniversalDV}. Prior work \cite{filippova2008dependency} implements with older dependency formalisms \cite{briscoe-etal-2006-second,Marneffe2006GeneratingTD}.} 
\item{In Table 1 of their original paper, \citet{filippova2013overcoming} provide an overview of the syntactic, structural, semantic and lexical features in their model. We implement every feature described in the table. We do not implement features which are not described in the paper.}
\item{\citet{filippova2013overcoming} augment edge labels in the dependency parse of $S$ as a preprocessing step. We reimplement this step using off-the-shelf augmented modifiers and augmented conjuncts available with the enhanced dependencies representation in CoreNLP \cite{Schuster2016EnhancedEU}.}
\item{\citet{filippova2013overcoming} preprocess dependency parses by adding an edge between the root node and all verbs in a sentence.\footnote{This step ensures that subclauses can be removed from parse trees, and then merged together to create a compression from different clauses of a sentence.} We found that replicating this transform literally (i.e. only adding edges from the original root to all tokens tagged as verbs) made it impossible for the ILP to recreate some gold compressions. (We suspect that this is due to differences in output from part-of-speech taggers). We thus add an edge between the root node and \textit{all} tokens in a sentence during preprocessing, allowing the ILP to always return the gold compression.}
\end{enumerate}

We assess convergence of the ILP by examining validation F1 score on the traditional sentence compression task. We terminate training after six epochs, when F1 score stabilizes (i.e. changes by fewer than $10^{-3}$ points).

\subsection{Implementation of SLOR}

We use the SLOR function to measure the readability of the shortened sentences produced by each compression system. SLOR normalizes the probability of a token sequence assigned from a language model by adjusting for both the probability of the individual unigrams in the sentence and for the sentence length.\footnote{Longer sentences are always less probable than shorter sentences; rarer words make a sequence less probable.} 

Following \cite{lau2015unsupervised}, we define the function as 

\begin{equation}
\text{SLOR}=\frac{\text{log}P_m(\xi) - \text{log}P_u(\xi)}{|\xi|}
\end{equation}

where $\xi$ is a sequence of words, $P_u(\xi)$ is the unigram probability of this sequence of words and $P_m(\xi)$ is the probability of the sequence, assigned by a language model.  $|\xi|$ is the length (in tokens) of the sentence.

We use a 3-gram language model trained on the training set of the \citet{filippova2013overcoming} corpus. We implement with KenLM \cite{Heafield-kenlm}. Because compression often results in shortenings where the first token is not capitalized (e.g.\ a compression which begins with the third token in $S$) we ignore case when calculating language model probabilities.

\subsection{Latency evaluation}
To measure latency, for each technique, we sample 100,000 sentences with replacement from the test set. We observe the mean time to compress each sentence using Python's built-in \textit{timeit} module. In order to minimize effects from unanticipated confounds in measuring latency, we repeat this experiment three separate times (with a one hour delay between experiments). Thus in total we collect 300,000 observations for each compression technique. We observe that runtimes are log normal, and thus report each latency as the geometric mean of 300,000 observations. We use an Intel Xeon processor with a clock rate of 2.80GHz.

\subsection{Compression ratios}

When comparing sentence compression systems, it is important to ensure that all approaches use the same rate of compression \cite{napoles2011evaluating}. Following \citet{filippova2015sentence}, we define the compression ratio as the character length of the compression divided by the character length of the sentence. We present test set compression ratios for all methods in Table \ref{t:cr}. Because ratios are similar, our comparison is appropriate.

\begin{table}[htb!]
\centering
\begin{tabular}{@{}l | l@{}}
\textsc{random} & 0.405\\ 
\textsc{ilp} &  0.408 \\
\textsc{ablated} & 0.387 \\
\textsc{vertex addition}$_{LR}$ &  0.403  \\
\textsc{vertex addition}$_{NN}$ &  0.405  \\ \midrule
$C_g$ Train  & 0.384 \\
$C_g$ Test   & 0.413 \\
\end{tabular}
\caption{Mean test time compression ratios for all techniques. We also show mean ratios for gold compressions $C_g$ across the train and test sets.}\label{t:cr}.
\end{table}


\end{document}